\documentclass{article}

    \PassOptionsToPackage{numbers, compress}{natbib}

    \usepackage[nonatbib, preprint]{neurips_2020}

\usepackage[utf8]{inputenc} 
\usepackage[T1]{fontenc}    
\usepackage{url}            
\usepackage{booktabs}       
\usepackage{amsfonts}       
\usepackage{nicefrac}       
\usepackage{microtype}      

\usepackage{xcolor}

\usepackage[ruled]{algorithm2e}

\usepackage{wrapfig}
\usepackage{amsmath}
\usepackage{amsfonts}
\usepackage{rotating} 

\DeclareMathOperator*{\Max}{max}

\newcommand{\KL}{D_{\mathrm{KL}}}

\title{Dynamics Generalization via Information Bottleneck in Deep Reinforcement Learning}

\author{%
  Xingyu Lu \\
  UC Berkeley\\ \And
  Kimin Lee  \\
  UC Berkeley\\ \And Pieter Abbeel  \\
  UC Berkeley\\ \And
  Stas Tiomkin  \\
  UC Berkeley
}

\begin{document}
\maketitle

\begin{abstract}
  Despite the significant progress of deep reinforcement learning (RL) in solving sequential decision making problems, RL agents often overfit to training environments and struggle to adapt to new, unseen environments. This prevents robust applications of RL in real world situations, where system dynamics may deviate wildly from the training settings. In this work, our primary contribution is to propose an information theoretic regularization objective and an annealing-based optimization method to achieve better generalization ability in RL agents. We demonstrate the extreme generalization benefits of our approach in different domains ranging from maze navigation to robotic tasks;  for the first time, we show that agents can generalize to test parameters more than 10 standard deviations away from the training parameter distribution. This work provides a principled way to improve generalization in RL by gradually removing information that is redundant for task-solving; it opens doors for the systematic study of generalization from training to extremely different testing settings, focusing on the established connections between information theory and machine learning.
\end{abstract}

\section{Introduction}
Dynamics generalization in deep reinforcement learning (RL) studies the problem of transferring a RL agent's policy from training environments to settings with unseen system dynamics or structures, such as the layout of a maze 
or the physical parameters of a robot \cite{nagabandi2018learning, lee2020context}. Although recent advancement in deep reinforcement learning has enabled agents to perform tasks in complex training environments, dynamics generalization remains a challenging problem \cite{rajeswaran2017towards, henderson2018deep}. 

Training policies that are robust to unseen environment dynamics has several merits. First and foremost, an agent trained in an ideal setting may be required to perform in more adversarial circumstances, such as increased obstacles, darker lighting and rougher surfaces. Secondly, it may enable efficient sim-to-real policy transfers \cite{tobin2017domain}, as the agent may quickly adapt to the differences in dynamics between the training environment and the testing environment. Lastly, an information bottleneck naturally divides a model into its encoder and controller components, improving the interpretability of end-to-end RL policies, which have traditionally been assumed as a black box. 


In this work, we consider the problem of dynamics generalization from an information theoretic perspective. Studies in the field of information bottleneck have shown that generalization of deep neural networks in supervised learning can be measured and improved by  controlling the amount of information flow between layers \cite{tishby2015deep}; in this paper, we hypothesize that the same can be applied to reinforcement learning.
In particular, we show that the poor generalization in unseen tasks is due to the DNNs memorizing environment observations, rather than extracting the relevant information for a task. To prevent this, we impose communication constraints as an information bottleneck between the agent and the environment. Such bottleneck would limit the information flow between observations and representations, thus encouraging the encoder to only extract relevant information from the environment and preventing memorization.

\begin{wrapfigure}[20]{r}{0.6\textwidth}
    \centering
    \includegraphics[width=0.6\textwidth]{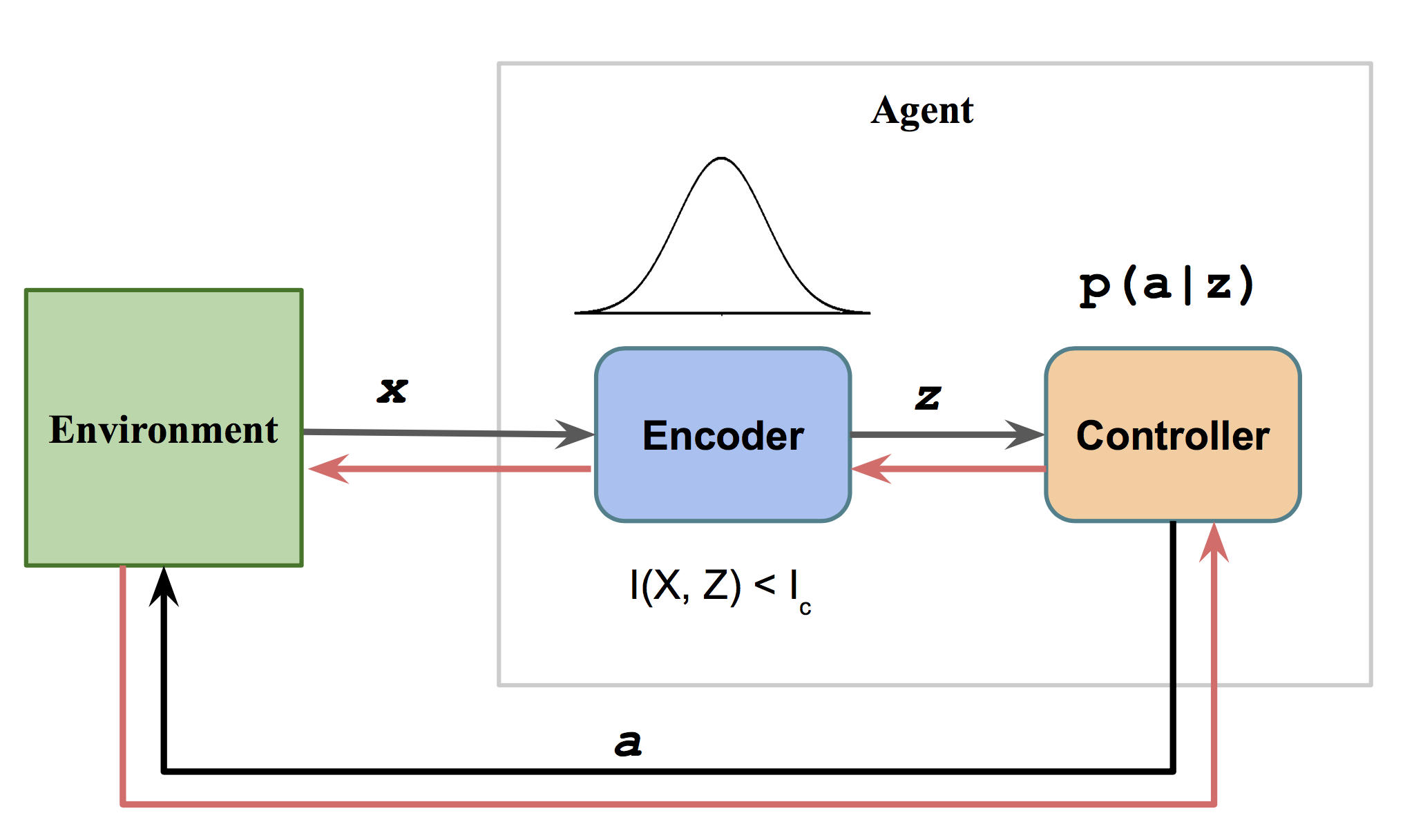}
    \caption{Illustration of the proposed scheme. We add stochasticity to the perception component (encoder) of the network, and constrain the information flow through it. The arrows in black indicate the input/output flow of each component, while the arrows in red indicate end-to-end gradients.}
    \label{img:blockplot}
\end{wrapfigure}

A joint optimisation of encoder and policy with an information bottleneck is a challenging problem, because, in general, the separation principle~\cite{witsenhausen1971separation} is not applicable. The separation principle allows to estimate state from observation (and under certain conditions to compress observation \cite{tanaka2017lqg}), and then to derive an  policy. In the cases where the separation principle is not applicable, a joint optimisation of encoder and policy can be seen as 'chicken and egg' problem: to derive an optimal policy one needs a meaningful state representation, which in turn depends on the performance of the policy. 

Our main contributions are as follows. Firstly, we tackle the problem of poor generalization of DRL to unseen tasks by applying an information bottleneck between observations and state representations (see Figure \ref{img:blockplot}). Specifically, we find a stochastic mapping from observations to internal representations, and regularize such mapping to limit the amount of information flow. Secondly and most significantly, we propose an annealing scheme for a stable join-optimization of the encoder and policy components of the network, finding a family of solutions parameterized by the weight of the information constraint. Thirdly, we demonstrate that policies trained with an information bottleneck achieve significantly better performance on tasks with unseen layouts, goals and dynamics, as compared to the standard DRL methods. Finally, we demonstrate that our method produces state representations which admit a semantic interpretation, which is in general not guaranteed for end-to-end DRL. Specifically, we demonstrate that the encoder in our approach maps stochastic observations to a space where distances between points are consistent with their values from the optimal critic.

Our proposed method is general and can be intergrated with most state-of-the-art reinforcement learning architectures. A version of our method based on a Pytorch baseline is published and available at github.com/anonymous.

\section{Related Work}
There is a series of previous works that address the problem of control with information bottlenecks. \cite{borkar1997lqg} is one of the first works in this direction, where the effects of state compression were studied in the case of linear and known dynamics. Specifically, they showed that in the case of Linear Quadratic Regulator, there exists an optimal compression scheme of state observations. The following works~\cite{tatikonda2004control, tatikonda2004stochastic, tanaka2017lqg, tiomkin2017unified}, studied the optimality of compression schemes under different assumptions, although all of them assumed known dynamics, and did not consider information bottleneck for its generalization benefits. 

Recently, it was shown that information bottleneck improves generalization in adversarial inverse reinforcement learning~\cite{peng2018variational}. By placing a bottleneck on the discriminator of a GAN, the author effectively balances the performance the discriminator and the generator to provide more meaningful gradients. This work, however, focuses strictly on imitation learning, and does not consider any online learning setting involving long-horizon planning.

Another relevant work is the work by Pacelli and Majumdar \cite{pacelli2020learning}, where information bottleneck is estimated and optimized through separate MINE estimators~\cite{belghazi2018mutual} at each time step. While this work also tackles the problem of generalization, it only focuses on image-based environments with changing textures, without considering changing environment goals or dynamics. Additionally, the use of separate MINE estimators at each time step may limit the scalability of the method for long horizon problems. Our work, in contrast, trains a single encoder whose information is regularized without any explicit estimators, and we focus on dynamics randomization problems with changing environment layouts and parameters.

Finally, in Goyal et al. \cite{goyal2019infobot}, the information bottleneck between actions and goals is studied with an aim to create goal independent policies. While both \cite{goyal2019infobot} and our work utilize the variational approximation of the upper bound on the mutual information, their work focuses on finding high information states for more efficient exploration, which is a different objective from our work.

\section{Preliminary}

\subsection{Markov Decision Process and Reinforcement Learning}
This paper assumes a finite-horizon Markov Decision Process (MDP) \cite{puterman1994markov}, defined by a tuple $(\mathcal{S}, \mathcal{A}, \mathcal{P}, r, \gamma, T)$. Here, $\mathcal{S} \in \mathbb{R}^d$ denotes the state space (which could either be noisy observations or raw internal states), $\mathcal{A} \in \mathbb{R}^m$ denotes the action space, $\mathcal{P} : \mathcal{S} \times \mathcal{A} \times \mathcal{S} \rightarrow \mathbb{R}^+$ denotes the state transition distribution, $r: \mathcal{S} \times \mathcal{A} \rightarrow \mathbb{R}$ denotes the reward function, ${\gamma \in [0, 1]}$ is the discount factor, and finally $T$ is the horizon. At each step $t$, the action $a_t \in \mathcal{A}$ is sampled from a policy distribution $\pi_\theta(a_t \vert s_t)$ where $s \in \mathcal{S}$ and ${\theta}$ is the policy parameter. After transiting into the next state {by sampling from} $p(s_{t+1} \vert a_t, s_t)$, where $p \in \mathcal{P}$, the agent receives a scalar reward $r(s_{t}, a_{t})$. {The agent continues performing actions until it enters a terminal state or $t$ reaches the horizon, by when the agent has completed one episode. We let $\tau$ denote the sequence of states that the agent enters in one episode.}

With such definition, the goal of RL is to learn a policy $\pi_{\theta^*}(a_t \vert s_t)$ that maximizes the expected discounted reward $\mathbb{E}_{\pi, P}[R(\tau_{0:T-1})] = \mathbb{E}_{\pi, P}[\sum_0^{T-1}\gamma^t r(s_{t}, a_{t})]${, where expectation is taken on the possible trajectories $\tau$ and the starting states $x_0$}. In this paper, we assume model-free learning, meaning the agent does not have access to the environment dynamics $\mathcal{P}$.

To study dynamics generalization, we further focus on context conditional environments, which correspond to a MDP distribution parameterized by a context variable $c$. Here $c$ could range from a robot's density to the coefficient of friction between any two surfaces. For each context $c$, the MDP adapts a specific state transition distribution $p_c(s'|s,a)$, and the agent now aims to learn a policy $\pi_{\theta^*}(a_t \vert s_t, c)$ that maximizes the reward given a particular context. Here, $c$ is directly provided to the agent as an oracle. Our goal is to train on a distribution of context $C_{train}$, and evaluate the agent's generalization performance on unseen contexts $c_{test} \notin C_{train}$.

\subsection{Mutual Information} 

Mutual information measures the amount of information obtained about one random variable after observing another random variable \cite{cover2012elements}. Formally, given two random variables $X$ and $Y$ with joint distribution $p(x, y)$ and marginal densities $p(x)$ and $p(y)$, their MI is defined as the KL-divergence between joint density and product of marginal densities:
\begin{align}
    MI(X;Y) = D_{KL}( p(x, y) \Vert p(x)p(y) ) = \mathbb{E}_{p(x, y)}[\textrm{log} \frac{p(x, y)}{p(x)p(y)}].
\end{align}

\section{Method}\label{sec:Method}
\subsection{Problem Definition}

We consider an architecture in which the agent learns with limited information from the environment: instead of learning directly from the environment states $s \in S$, the agent needs to estimate noisy encoding $z \in Z$ of the state, whose information is limited by a bottleneck.

Formally, we decompose the agent policy $\pi_{\theta}$ into an encoder $f_{\theta_1}$ and a decoder  $g_{\theta_2}$ (action policy), where $\theta = \{\theta_1, \theta_2\}$. The encoder maps environment states into stochastic embedding, and the decoder outputs agent actions $a \in A$:
\begin{align}
p_{\pi_{\theta}}(a \vert s) &= \int_{z} p_{g_{\theta_2}}(a \vert z) p_{f_{\theta_1}}(z \vert s) dz 
\end{align}
With such setup, we maximize the RL objective with a constraint on the mutual information between the environment states and the embedding:
\begin{align}
    J(\theta) = \Max_{\theta}& \quad \mathbb{E}_{\pi_{\theta}, \tau} [R(\tau)], \quad s.t. \quad I(Z, S) \leq I_{c}
\end{align}

To estimate mutual information between $S$, and $Z$, We makes use of the following identity:
\begin{equation}
    I(Z, S) = \KL\,[p(Z, S) \,\vert\, p(Z)p(s)] = \mathbb{E}_{S}\,[ \KL\,[p(Z \vert S) \,\vert \, p(Z)]] 
\end{equation}

In practice, we take samples of $\KL\,[p(Z \vert S) \,\vert \, p(Z)]$ to estimate the mutual information. While $p(Z \vert S)$ is straightforward to compute, calculating $p(Z)$ requires marginalization across the entire state space $S$, which in most non-trivial environments are intractable. Instead, we follow the method adopted in many recent works and introduce an approximator, $q(Z) \sim \mathcal{N}(\Vec{0}, \mathbb{I})$ , to replace $p(Z)$ \cite{peng2018variational, goyal2019infobot}. A proof for this can be found in the Appendix.

\subsection{Unconstrained Lagrangian}
We introduce a Lagrangian multiplier $\beta$ and optimize on the upper bound of $I(Z, S)$ given by the approximator $q(Z)$:
\begin{equation}
    \mathcal{L}(\theta) = \Max_{\theta}\; \mathbb{E}_{\pi_{\theta}, \tau} [R(\tau)] - \beta \mathbb{E}_{S}[\KL\,[p(Z \vert S) \, \vert \, q(Z)]]
\end{equation}

\begin{figure*}

\centering
  \begin{minipage}{.25\textwidth}
      \centering
  \begin{minipage}{0.9\textwidth}
    \includegraphics[width=\linewidth]{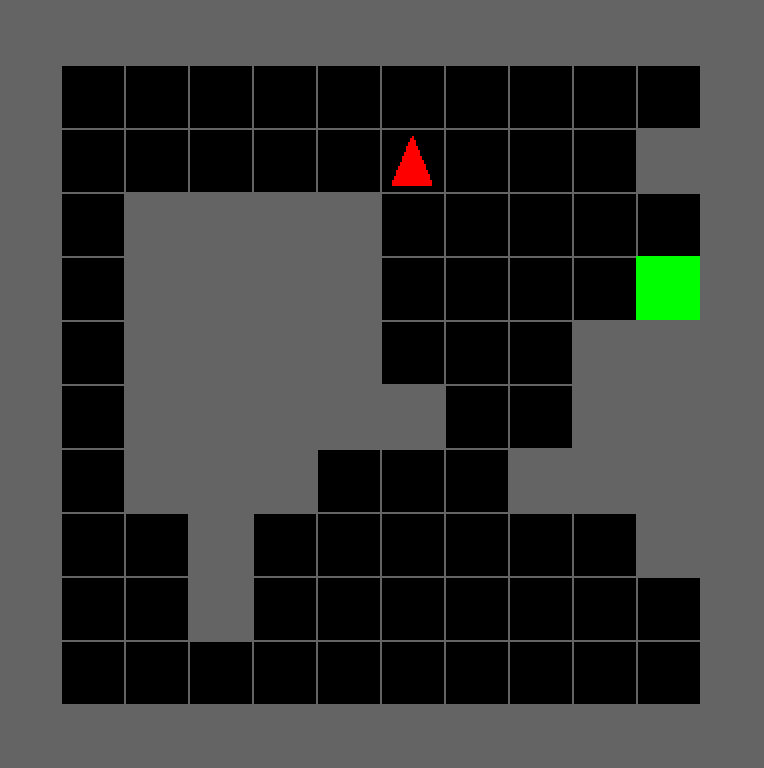}
  \end{minipage}%
    \end{minipage}%
  \begin{minipage}{.25\textwidth}
      \centering
  \begin{minipage}{0.9\textwidth}
    \includegraphics[width=\linewidth]{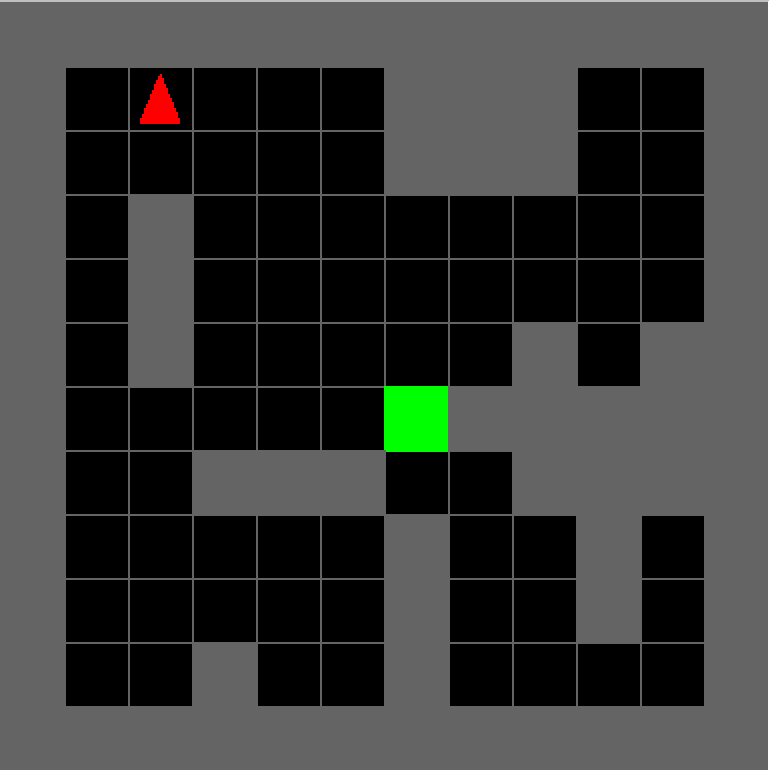}
  \end{minipage}%
    \end{minipage}%
  \begin{minipage}{.25\textwidth}
      \centering
  \begin{minipage}{0.9\textwidth}
    \includegraphics[width=\linewidth]{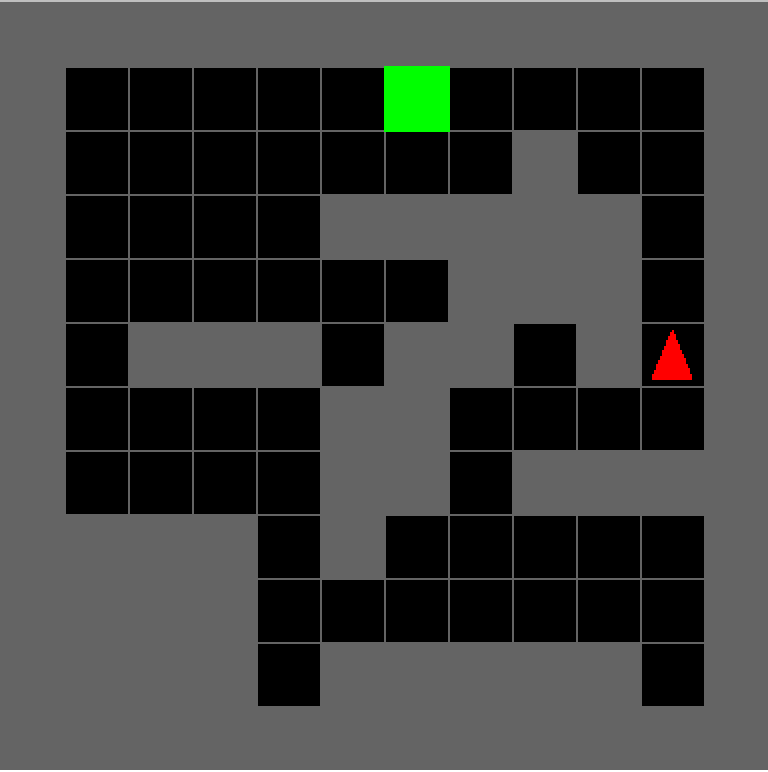}
      \end{minipage}%
  \end{minipage}%
    \caption{Visualization of examples of grid layouts used in this paper, sampled from randomly generated layouts.}
    \label{img:mazes}
\end{figure*}

As discussed in \cite{strouse2018learning}, the gradient update at time $t$ is the policy gradient update with the modified reward, minus a scaled penalty by KL-divergence between state and embedding:
\begin{align}
    \nabla_{\theta, t} \mathcal{L}(\theta) =R'(t)\nabla_{\theta}\log(\pi_{\theta}(a_{t}, s_{t})) \nonumber - \beta\nabla_{\theta} \KL[p(Z \vert s) \, \vert \, q(Z)] \label{eq:update}
\end{align}
where $R'(t) = \sum_{i =1}^{t} \gamma^{i}r'(a_t, s_t)$ is the discounted reward until step $t$, and $r'(a_t, s_t)$ is the environment reward $r(a_t, s_t)$ modified by the KL penalty: $r'(a_t, s_t) = r(a_t, s_t) + \beta\KL[p(Z \vert s) \, \vert \, q(Z)]$.

\subsection{Annealing Scheme} \label{sec:anneal}

We generate a family of solutions (optimal pairs of encoder and policy) parametrized by the information bottleneck constraint weight $\beta$. In our case, each solution is characterized by a correspondingly constrained amount of information required to maximize the environment rewards. 

The rationale is as follows: to encourage the agent to extract relevant information from the environment, we want to impose high penalty for passing too much information through the encoder. At the beginning of training, such penalty produces gradients that offsets the agent's learning gradients, making it difficult for the agent to form good policies.

To tackle this problem, we create the entire family of solutions through \textit{annealing}, starting from a deterministic (unconstrained) encoder, and gradually injecting noise by increasing the penalty coefficient (\textit{temperature parameter}), $\beta$.

\begin{wrapfigure}[19]{r}{0.55\textwidth}
    \centering
    \includegraphics[width=0.55\textwidth]{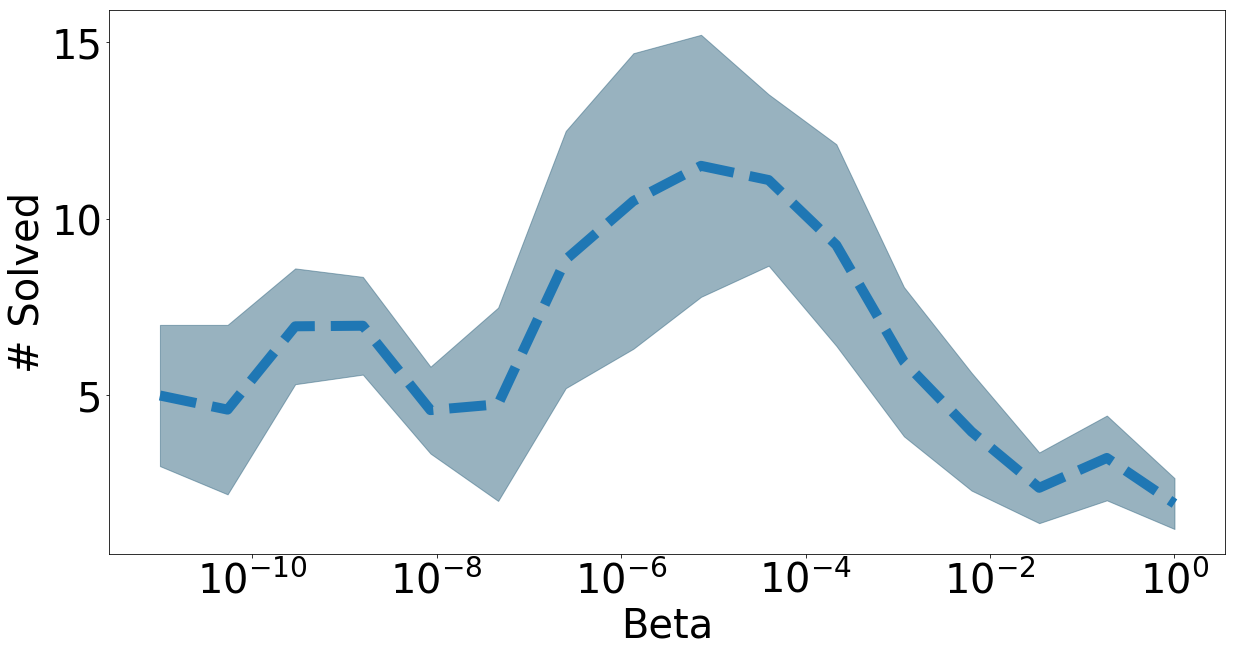}
    \caption{The number of successful configurations (reward > 150) out of 20 unseen test configurations for the Cartpole environment at different beta values along the annealing curve. An increase in generalization ability is observed between $10^{-7} < \beta < 10^{-5}$, followed by a sharp drop in generalization ability.}
    \label{img:annealcp}
\end{wrapfigure}

This approach allows training of well-formed policies for much larger $\beta$ values, as the encoder has already learned to extract useful information from the environment, and only needs to learn to "forget" more information as $\beta$ increases. In the experiment section, we will demonstrate that training the model using annealing enables the agent to learn with much larger $\beta$ coefficients compared to from scratch. In particular, Figure \ref{img:annealcp} shows an increase and decrease in generalization benefits along the annealing curve.

\section{Experiment Results}\label{sec:Experiments}
In this section, we apply the approaches described in Section \ref{sec:Method} to discrete maze environments and various control environments. In doing so, we aim to answer the following questions:
\begin{enumerate}
    \item How effectively can we learn a policy with information bottleneck through annealing?
    \item How well can a policy trained end-to-end with an information bottleneck transfer to new, unseen structure or dynamics?
\end{enumerate}


\subsection{Mazes}

\begin{figure*}[h]
    \centering
  \begin{minipage}{.45\textwidth}
      \centering
  \begin{minipage}{0.9\textwidth}
    \includegraphics[width=\linewidth]{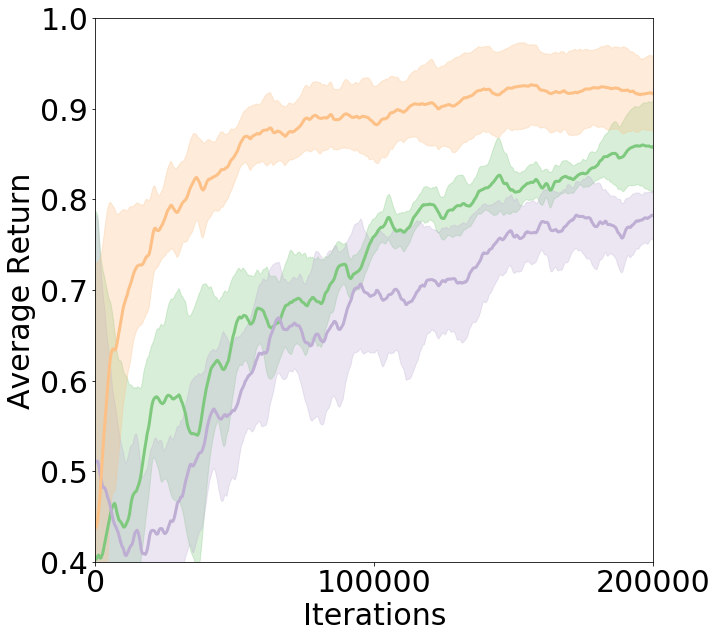}
  \end{minipage}%
    \end{minipage}%
  \begin{minipage}{.45\textwidth}
      \centering
  \begin{minipage}{1\textwidth}
    \includegraphics[width=\linewidth]{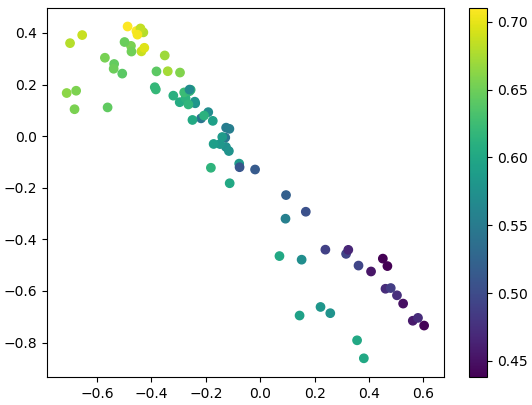}
  \end{minipage}%
    \end{minipage}%
  
    \caption{Learning curves of randomly generated mazes for baseline and different information bottlenecks (left); T-SNE projection of the encoder output for every state on 2D plane (right). The orange curve reaches near-optimal values the fastest, while the green and purple curves are very similar. For the T-SNE plot, there exist 1) a consistent color gradient along the diagonal by critic values 2) branching by optimal actions. }
    \label{img:TSNE}
    \label{img:mazecompare}
\end{figure*}



MiniGrid Environments are used as the primary discrete experiments \cite{gym_minigrid}. To validate the results statistically, we randomly generate and sample maze environments of the same size to test the agent's ability to transfer to new layouts. The fixed layout and examples of the randomly generated layouts are listed in Figure \ref{img:mazes}. 

\begin{figure*}[b]
  \begin{minipage}{.32\textwidth}
      \centering
  \begin{minipage}{0.9\textwidth}
    \includegraphics[width=\linewidth]{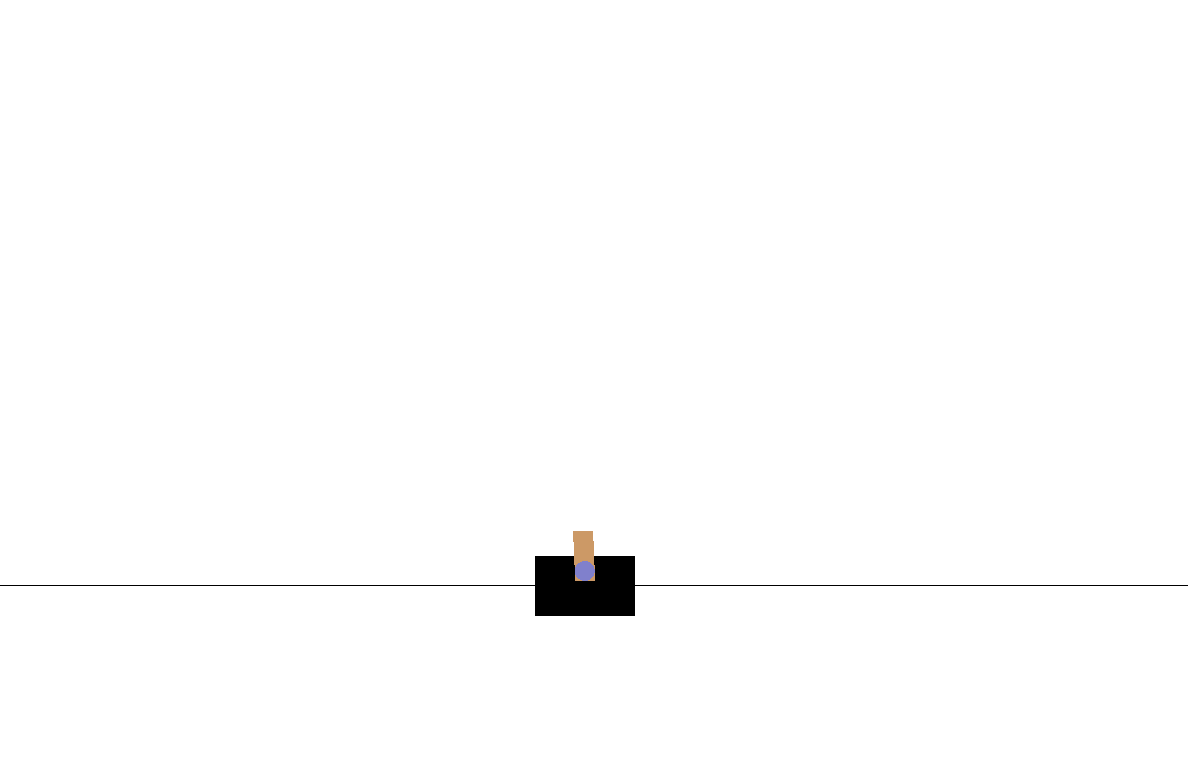}
  \end{minipage}%
    \end{minipage}%
  \begin{minipage}{.32\textwidth}
      \centering
  \begin{minipage}{0.9\textwidth}
    \includegraphics[width=\linewidth]{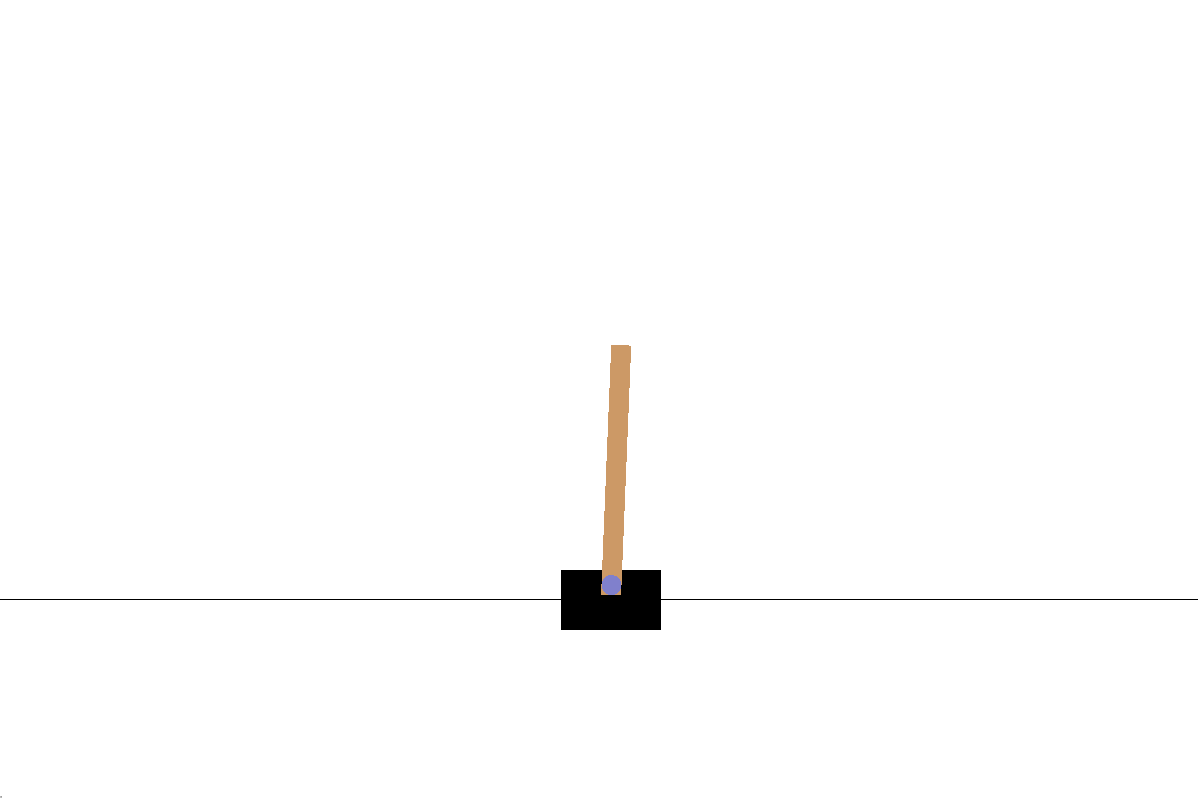}
  \end{minipage}%
    \end{minipage}%
  \begin{minipage}{.32\textwidth}
      \centering
  \begin{minipage}{0.9\textwidth}
    \includegraphics[width=\linewidth]{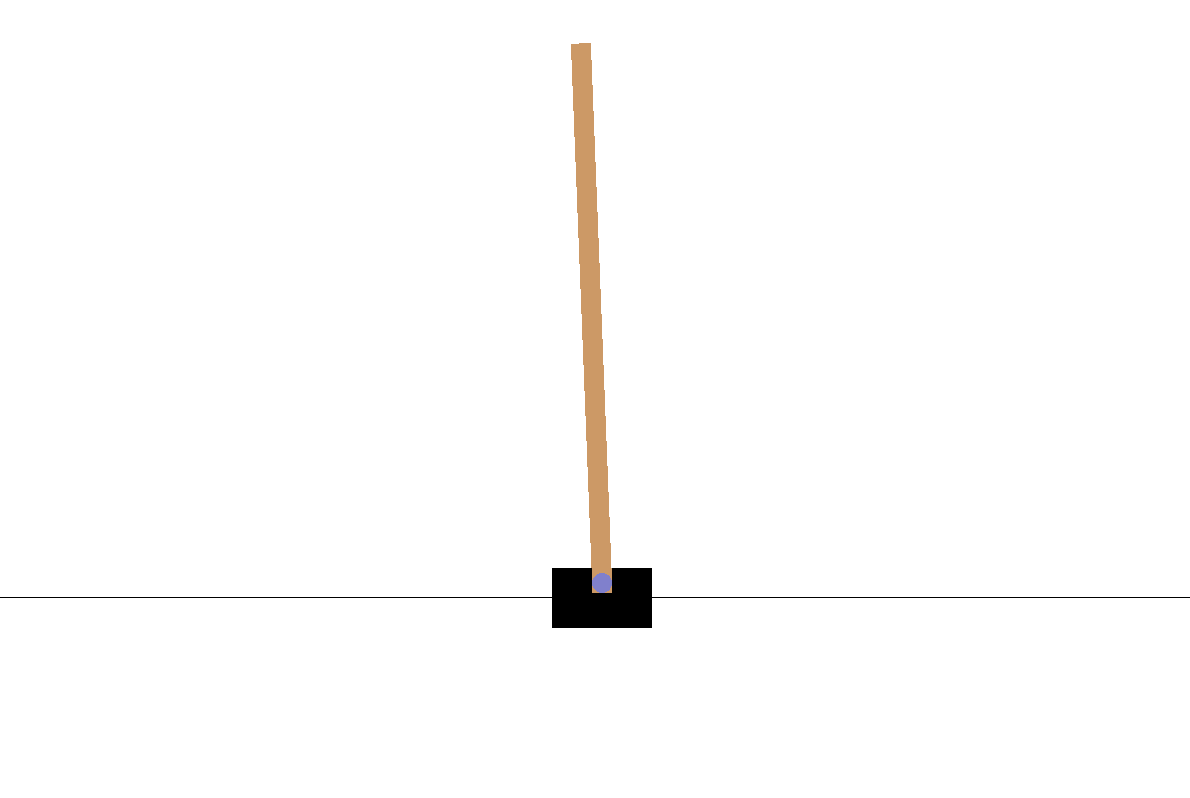}
  \end{minipage}%
    \end{minipage}%
  
    \caption{Visualization of 3 different pole lengths in the CartPole environment. The lengths are: 0.1 (left), 0.5 (middle), and 1.3 (right). The middle configuration is included in training, while the configurations on two sides are seen only during testing.}
    \label{img:cartpole}
\end{figure*}

For each transfer experiment, we randomly sample 4 mazes, 3 of which are used for the training set and 1 for testing. Specifically, we train a policy using the training set, then retain it for the unseen maze to assess how fast the model learns the new maze layout. Figure \ref{img:mazecompare} shows the learning curves of three different setups: learning with a tight information bottleneck ($\beta = 0.05$); learning with a loose information bottleneck ($\beta = 0.0001$ as \textit{ablation}); learning with full information ($\beta = 0$ and deterministic encoder as \textit{baseline}). As the plot shows, learning with a tight information bottleneck achieves the best transfer learning result, reaching near-optimal solution of 0.9 mean reward around 2 times faster compared to the baseline. The close performance between the baseline and the ablation suggests the benefit of generalization only emerges as we tighten the information bottleneck.

Furthermore, we demonstrate that the code learned through information bottleneck learns structured information about the maze. Figure \ref{img:TSNE} illustrates the projection of every state's embedding (after convergence) onto 2D space through T-SNE, with each point colored by its critic value. From the projection plot, we observe the emergence of consistent value gradients as well as local clustering by actions.

\subsection{CartPole}

\begin{figure*}[h]
\centering
  \begin{minipage}{.5\textwidth}
      \centering
  \begin{minipage}{1\textwidth}
    \includegraphics[width=\linewidth]{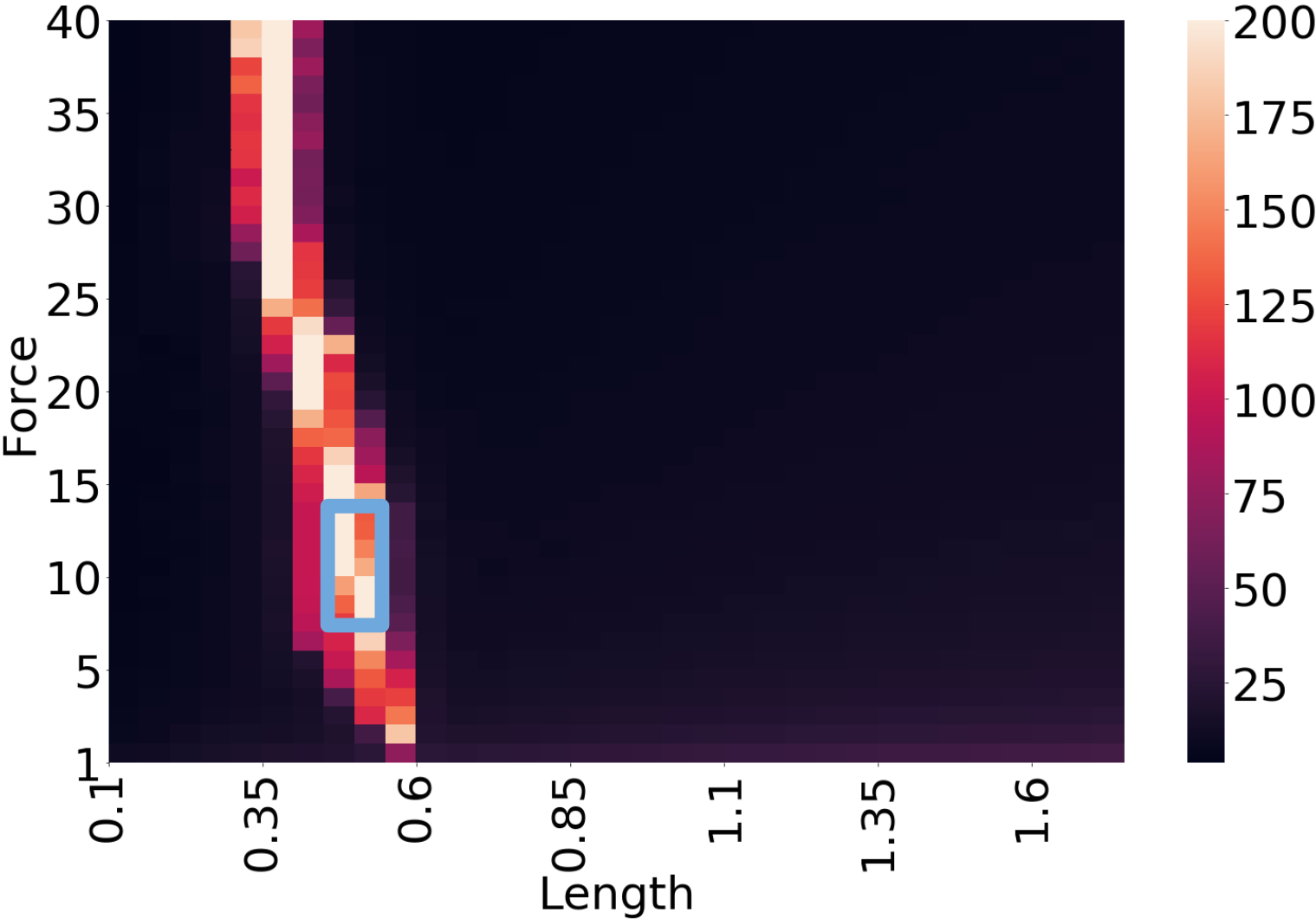}
  \end{minipage}%
    \end{minipage}%
  \begin{minipage}{.5\textwidth}
      \centering
  \begin{minipage}{1\textwidth}
    \includegraphics[width=\linewidth]{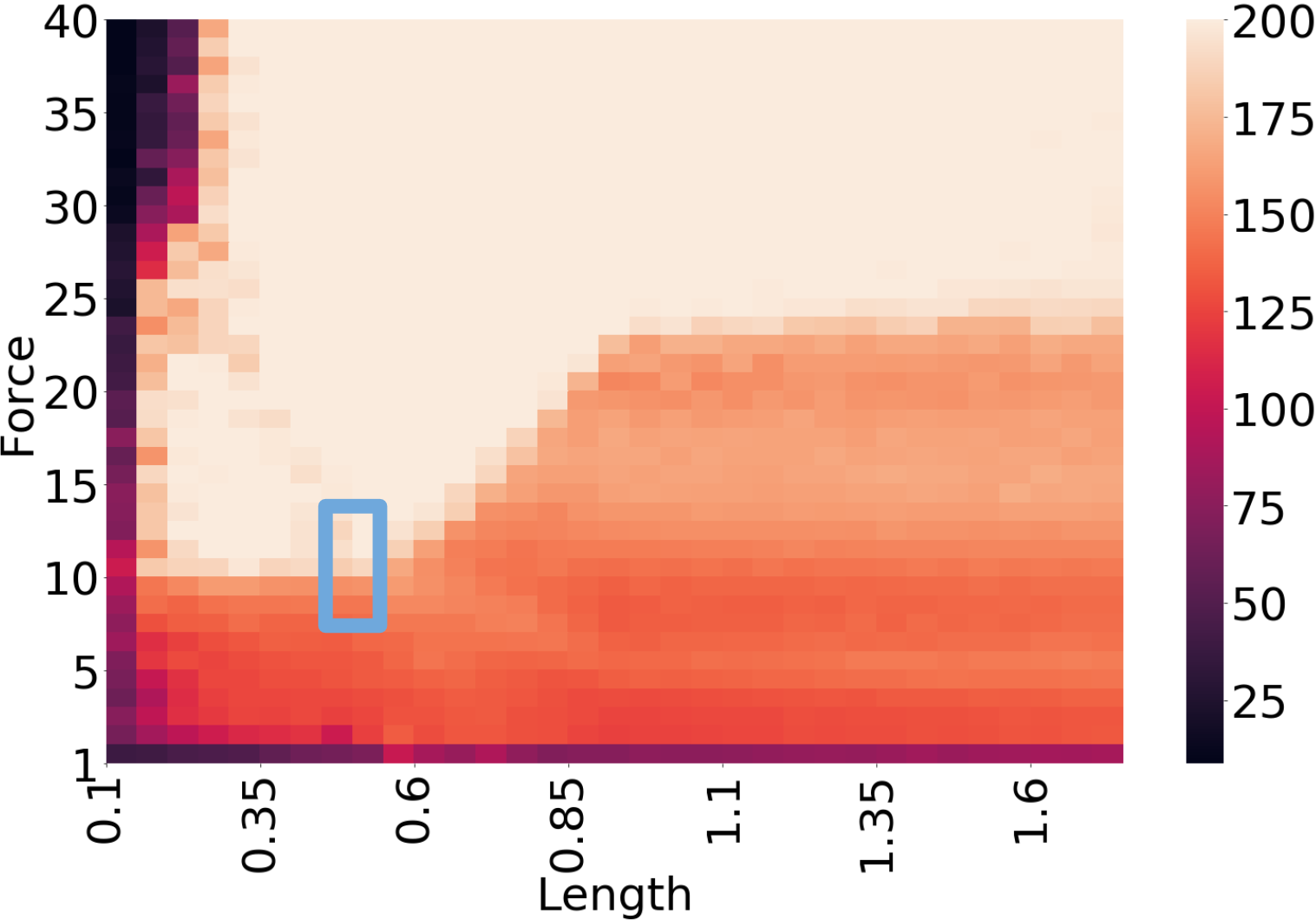}
  \end{minipage}%
    \end{minipage}%
  
    \caption{Evaluation performance of policies trained through baseline (left) and information bottleneck (right) on CartPole. The x-axis indicates the length of the pole, while the y-axis indicates the push force of the cart. Each evaluation result is averaged over 20 episodes. The training set is boxed.}
    \label{img:cp_eval}
\end{figure*}
The CartPole environment consists of a pole attached to a cart sliding on a frictionless surface. The pole is free to swing around the connection point to the cart, and the environment goal is to move the cart either left or right to keep the pole upright. The agent obtains a reward of 1 for keeping the pole upright at each time step, and can achieve a maximum of $200$ reward over the entire episode. Should the pole fail to maintain an angle of 12 degrees from the vertical line, the episode will terminate early.

The CartPole environment is configured to have 2 discrete actions: moving left or right at each time step. For this environment, we vary two environment parameters: the magnitude of the cart's push force, and the length of the pole. The push force affects the cart's movement at each time step, while the length of the pole affects its torque. We provide limited randomization during training compared to the configurations in \cite{packer2018assessing}: we range push forces from $7$ to $13$, and the pole length from $0.45$ to $0.55$. For evaluation, we consider a much wider range as well as extreme values: we first test the policy's performance on push forces ranging from $1$ to $40$ and pole lengths from $0.1$ to $1.7$; then, we test on extremely large values of push forces ($80$, $160$) and pole lengths ($1.7$, $3.4$, $6.8$) to assess the policy's stability. While push force is difficult to visualize, Figure \ref{img:cartpole} illustrates the different pole lengths used for training and evaluation.



As illustrated in Figure \ref{img:cp_eval}, both the baseline and our approach achieve good training performance; the baseline, however, fails to generalize beyond unseen pole lengths, while our method produces a policy that adapts to almost all test configurations. The difference in generalization to unseen dynamics between the baseline and our approach showcases the power of information bottleneck: by limiting the amount of information flow between observation and representation, we force the DNN to learn a general representation of the environment dynamics that can be readily adapted to unseen values.

A policy trained with a well-tuned bottleneck performs well even in extreme configurations. For the extreme ranges (force $\in \{80, 160\}$ and pole length $\in \{1.7, 3.4, 6.8\}$), the agent trained with a bottleneck achieves optimal reward (> 195) on all configurations. The plot for this result is moved to the Appendix.

\begin{figure*}[t]
    \centering
  \begin{minipage}{.35\textwidth}
      \centering
  \begin{minipage}{0.9\textwidth}
    \includegraphics[width=\linewidth]{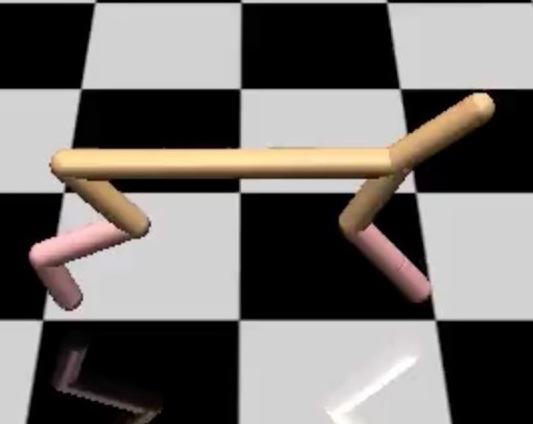}
  \end{minipage}%
    \end{minipage}%
  \begin{minipage}{.35\textwidth}
      \centering
  \begin{minipage}{0.9\textwidth}
    \includegraphics[width=\linewidth]{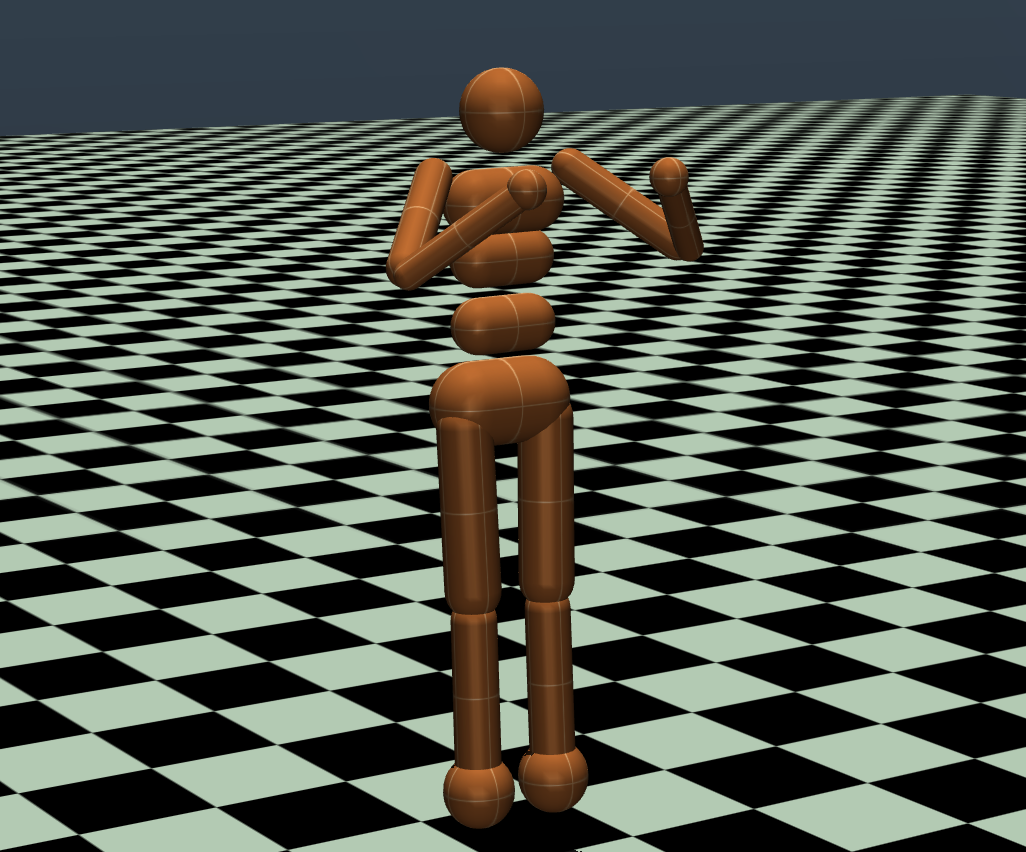}
  \end{minipage}%
    \end{minipage}%
  
    \caption{Visualization of the HalfCheetah and Humanoid environments. Although not visually different, every configuration of each robot corresponds to different physical parameters that alter their movement dynamics. }
    \label{img:hcsh}
\end{figure*}

\begin{wrapfigure}[27]{r}{0.5\textwidth}
    \centering
    \includegraphics[width=0.5\textwidth]{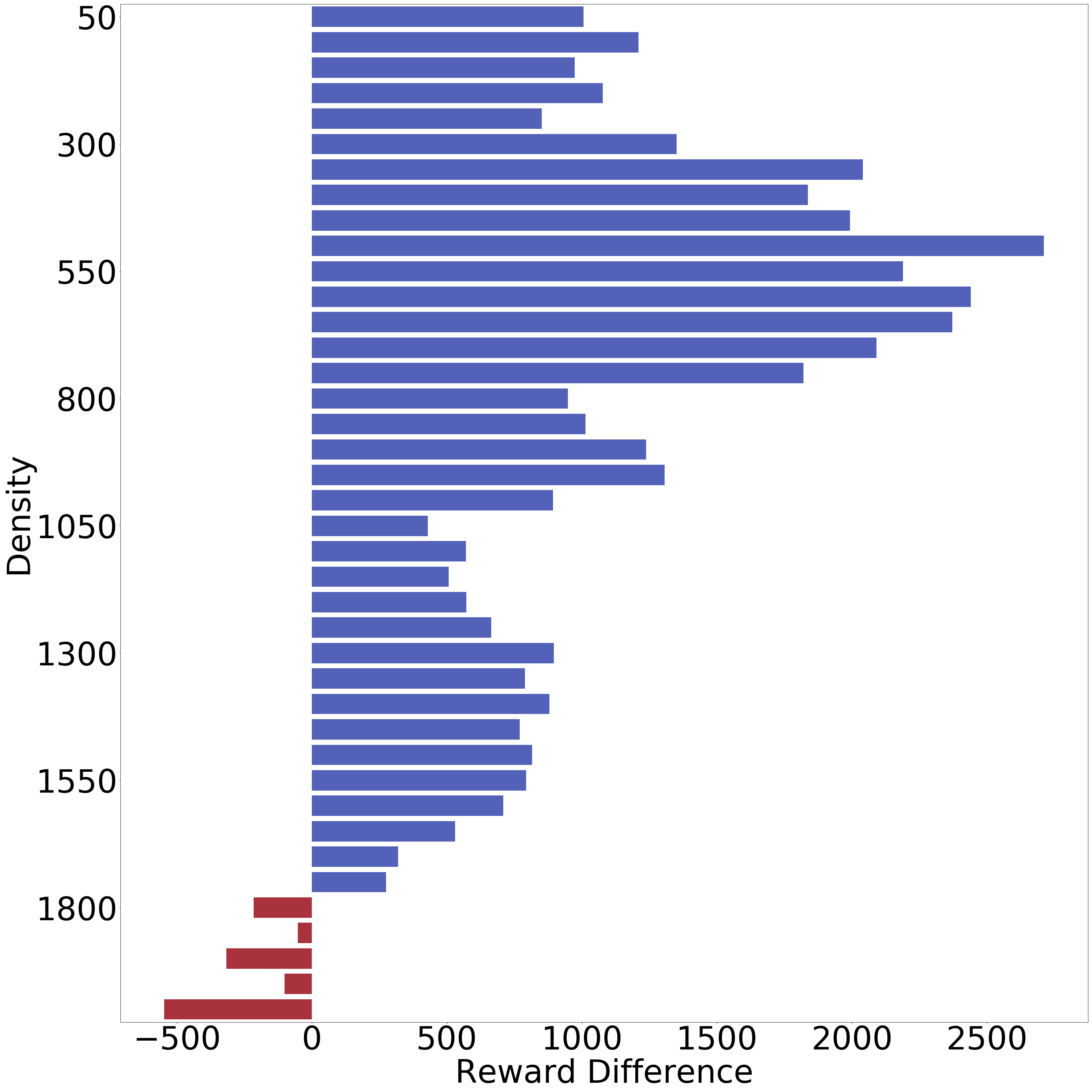}
    \caption{Visualization of the reward difference between baseline and our method. The y-axis indicates the torso density, while the x-axis indicates the reward difference averaged over 20 episodes. The red bars indicate configurations where baseline achieves higher reward, and the blue bars indicate where our method performs better.}
    \label{img:hcdiff}
\end{wrapfigure}

\subsection{HalfCheetah}

Next, we demonstrate the generalization benefits of our method in the HalfCheetah environment. In this environment, a bipedal robot with 6 joints and 8 links imitates a 2D cheetah, and its goal is to learn to move in the positive direction without falling over. The environment reward is a combination of its velocity in the positive direction and the cost of its movement (in the form of a L-2 cost on action). A illustration of the environment is provided in Figure \ref{img:hcsh}.

The environment has continuous actions corresponding to the  force values applied to its joints. Its dynamics is more complex in nature compared to CartPole, making generalization a challenging task. Similar to \cite{packer2018assessing}, we vary the torso density of the robot to change its movement dynamics. In particular, we vary the training density from $750$ to $1000$, and test the policy's performance on density values ranging from $50$ to $2000$. As the robot's actions corresponding to forces, whose effects are linearly affected by density, policy extrapolation from the training parameters to the test parameters is extremely challenging.

While both the baseline's and our method's performances suffer outside of the training range, our method achieves significantly better reward when the density is low. Figure \ref{img:hcdiff} better illustrates the performance difference between the baseline and our method: for most test configurations our method performs significantly better than the baseline, especially for density values that are lower than those seen in testing. This again indicates better stability and generalization in the policy trained with an information bottleneck.

\subsection{Humanoid}


Finally, in the Humanoid environment (Figure \ref{img:hcsh}) a human-like robot with 13 rigid links and 17 actuators freely moves on a flat surface. The goal is to move forward as soon as possible, while keeping the cost of action low. The environment reward is the forward velocity of the center of the robot minus a L-2 penalty on the action.

Similar to HalfCheetah, the environment has continuous actions corresponding to the force values applied to the robot's joints. Another challenging environment, Humanoid tests a policy's ability to generalize a high dimensional system. For our experiments, we scale both the robot's mass and its joints' damping factors from $0.8$ to $1.25$, then testing the policy's performance on test mass and damping scales from $0.5$ to $1.55$. Both of these parameters directly affect the robot's actions' impact on movement.

\begin{wrapfigure}[19]{l}{0.6\textwidth}
    \centering
    \includegraphics[width=0.6\textwidth]{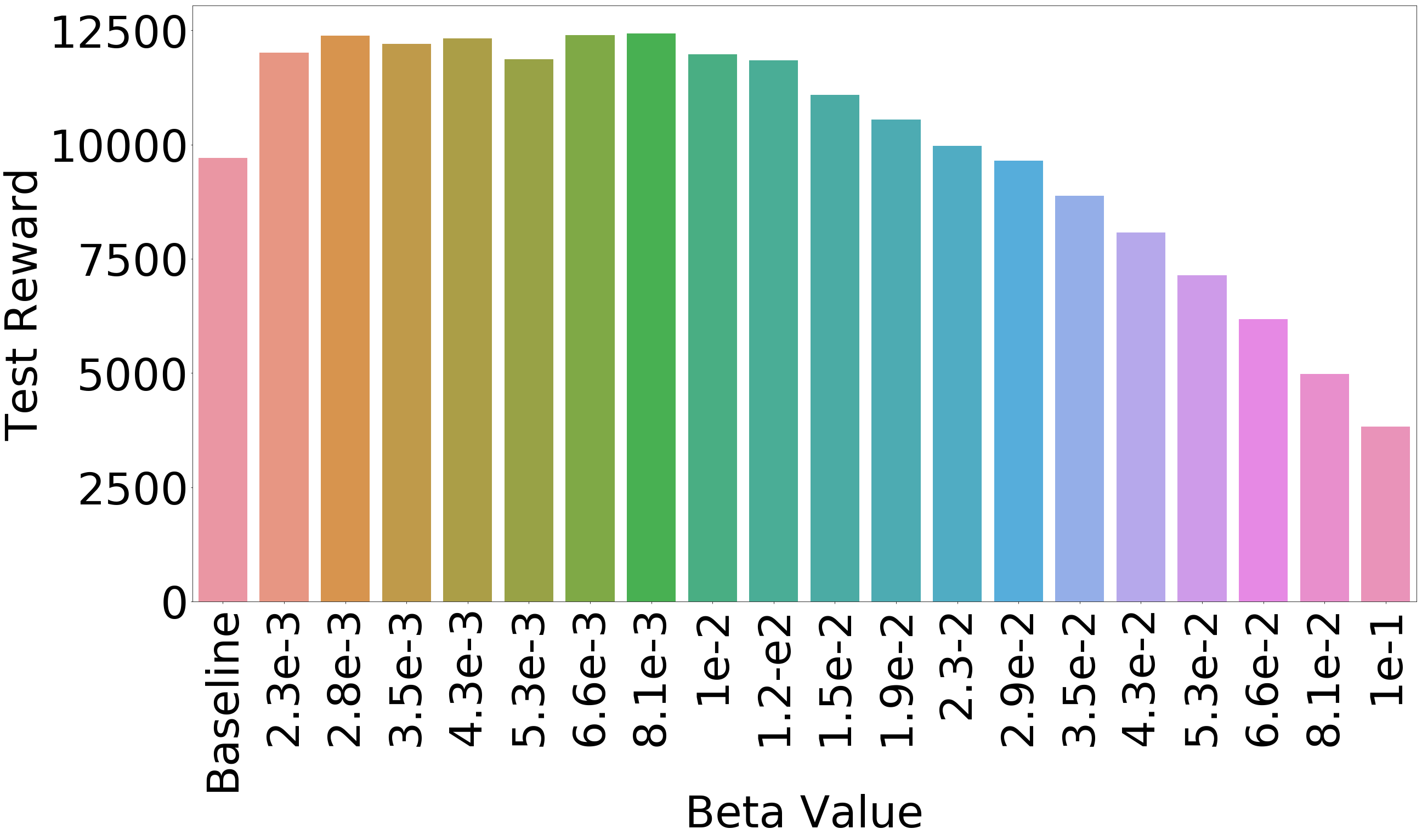}
    \caption{Visualization of the average evaluation reward of the baseline and our method in Humanoid. The x-axis indicates the beta value or the baseline, while the y-axis is the average test reward on unseen configurations only.}
    \label{img:shbars}
\end{wrapfigure}

The result for Humanoid is presented in Figure \ref{img:shbars}, where average test reward (on unseen parameters only) along the different beta values are shown alongside the baseline reward. In particular, for a properly tuned bottleneck, our method achieves significantly better performance than the baseline: for a $\beta$ value of $8e-3$, the average test reward is around 30\% higher than that of the baseline's average test reward, signifying a substantial boost in generalization performance. 

\section{Conclusion and Future Work}
In this work we proposed a principled way to improve generalization to unseen tasks in deep reinforcement learning, by introducing a stochastic encoder with an information bottleneck optimized through annealing.

We have proved our hypothesis that generalization in DRL can be improved by preventing explicit memorization of training environment observations. We showed that an explicit information bottleneck in the DRL cascade forces the agent to learn to squeeze the minimum amount of information from the observation before the optimal solution is found, preventing it from overfitting onto the training tasks. This led to much better generalization performances (for unseen maze layouts, unseen goals, and unseen dynamics) than baselines and other regularization techniques such as L-2 penalty and dropout. 

Practically, we showed that the suggested annealing scheme allowed the agent to find optimal encoder-decoder pairs under different information constraints, even for significant information compression that corresponds to very large $\beta$ values. This annealing scheme was designed to gradually inject noise to the encoder to reduce information (by gradually increasing $\beta$), while keeping a well-formed decoder (action policy) that received meaningful RL gradients. This slow change in the values of $\beta$ is critical, when it is not guarantied to have an optimal joint solution for the encoder and decoder (action policy), as in cases where the separation principle is not satisfied. 

Overall, we found significant generalization advantages of our approach over the baseline in the maze environment as well as control environment such as CartPole, HalfCheetah, and Humanoid. A CartPole policy trained using an information bottleneck, for instance, was able to generalize to test parameters more than 10 times larger than the training parameters, completely beating the baseline's generalization performance.

A promising future direction for research is to rigorously study the properties of the representation space, which may contribute to improving the interpretability of representations in deep neural networks in general. One of the insights of this work was that the produced representation in the maze environments preserved critic value distances of the original states; the representation space was thus consistent with the planning space, allowing generalization over unseen layouts.
\clearpage
\section{Broader Impact}
Our work improves the generalization ability of RL agents to extreme unseen environment dynamics, and can contribute to current efforts to deploy RL agents in real world circumstances. For instance, applying our method to an autonomous vehicle may boost its ability to navigate in extreme weather conditions, improving its safety for passengers;  a household robot (e.g. a laundry-folding robot) may better serve people by adapting to variations in its task due to the complex nature of the real world; production robots may operate more efficiently by better handling misplaced materials or components. As our method is general and can be plugged into any RL architectures, it can be potentially employed in existing systems to further boost their ability to handle edge cases in their tasks. 

While adding stochasticity to the system is common in reinforcement learning \cite{haarnoja2018soft, ppo, haarnoja2017reinforcement}, our method’s focus on injecting noise into the agent may cause it to operate falsely in rare occasions, due to the noisy encoder producing outlier codes. Thus, while we have demonstrated that on expectation our method achieves good generalization performance in extreme test settings, further studies in this direction with worst case optimality guarantees in mind are required. One possibility is to decrease significantly stochasticity in the encoder during test time, which may decrease performance but will prevent outlier codes; another potential direction is to consider empowerment or other metrics as safety measures to prevent the agent from taking extreme actions.

\section{Acknowledgement}
This work was supported in part by NSF under grant NRI-\#1734633 and by Berkeley Deep Drive. 
\bibliographystyle{unsrt}
\bibliography{references}

\begin{thebibliography}{10}

\bibitem{nagabandi2018learning}
Anusha Nagabandi, Ignasi Clavera, Simin Liu, Ronald~S Fearing, Pieter Abbeel,
  Sergey Levine, and Chelsea Finn.
\newblock Learning to adapt in dynamic, real-world environments through
  meta-reinforcement learning.
\newblock {\em arXiv preprint arXiv:1803.11347}, 2018.

\bibitem{lee2020context}
Kimin Lee, Younggyo Seo, Seunghyun Lee, Honglak Lee, and Jinwoo Shin.
\newblock Context-aware dynamics model for generalization in model-based
  reinforcement learning.
\newblock In {\em ICML}, 2020.

\bibitem{rajeswaran2017towards}
Aravind Rajeswaran, Kendall Lowrey, Emanuel~V Todorov, and Sham~M Kakade.
\newblock Towards generalization and simplicity in continuous control.
\newblock In {\em Advances in Neural Information Processing Systems}, pages
  6550--6561, 2017.

\bibitem{henderson2018deep}
Peter Henderson, Riashat Islam, Philip Bachman, Joelle Pineau, Doina Precup,
  and David Meger.
\newblock Deep reinforcement learning that matters.
\newblock In {\em Thirty-Second AAAI Conference on Artificial Intelligence},
  2018.

\bibitem{tobin2017domain}
Josh Tobin, Rachel Fong, Alex Ray, Jonas Schneider, Wojciech Zaremba, and
  Pieter Abbeel.
\newblock Domain randomization for transferring deep neural networks from
  simulation to the real world.
\newblock In {\em 2017 IEEE/RSJ international conference on intelligent robots
  and systems (IROS)}, pages 23--30. IEEE, 2017.

\bibitem{tishby2015deep}
Naftali Tishby and Noga Zaslavsky.
\newblock Deep learning and the information bottleneck principle.
\newblock In {\em 2015 IEEE Information Theory Workshop (ITW)}, pages 1--5.
  IEEE, 2015.

\bibitem{witsenhausen1971separation}
Hans~S Witsenhausen.
\newblock Separation of estimation and control for discrete time systems.
\newblock {\em Proceedings of the IEEE}, 59(11):1557--1566, 1971.

\bibitem{tanaka2017lqg}
Takashi Tanaka, Peyman~Mohajerin Esfahani, and Sanjoy~K Mitter.
\newblock Lqg control with minimum directed information: Semidefinite
  programming approach.
\newblock {\em IEEE Transactions on Automatic Control}, 63(1):37--52, 2017.

\bibitem{borkar1997lqg}
Vivek~S Borkar and Sanjoy~K Mitter.
\newblock Lqg control with communication constraints.
\newblock In {\em Communications, Computation, Control, and Signal Processing},
  pages 365--373. Springer, 1997.

\bibitem{tatikonda2004control}
Sekhar Tatikonda and Sanjoy Mitter.
\newblock Control under communication constraints.
\newblock {\em IEEE Transactions on automatic control}, 49(7):1056--1068, 2004.

\bibitem{tatikonda2004stochastic}
Sekhar Tatikonda, Anant Sahai, and Sanjoy Mitter.
\newblock Stochastic linear control over a communication channel.
\newblock {\em IEEE transactions on Automatic Control}, 49(9):1549--1561, 2004.

\bibitem{tiomkin2017unified}
Stas Tiomkin and Naftali Tishby.
\newblock A unified bellman equation for causal information and value in markov
  decision processes.
\newblock {\em arXiv preprint arXiv:1703.01585}, 2017.

\bibitem{peng2018variational}
Xue~Bin Peng, Angjoo Kanazawa, Sam Toyer, Pieter Abbeel, and Sergey Levine.
\newblock Variational discriminator bottleneck: Improving imitation learning,
  inverse rl, and gans by constraining information flow.
\newblock {\em ICLR 2019}, 2019.

\bibitem{pacelli2020learning}
Vincent Pacelli and Anirudha Majumdar.
\newblock Learning task-driven control policies via information bottlenecks.
\newblock {\em arXiv preprint arXiv:2002.01428}, 2020.

\bibitem{belghazi2018mutual}
Mohamed~Ishmael Belghazi, Aristide Baratin, Sai Rajeshwar, Sherjil Ozair,
  Yoshua Bengio, Devon Hjelm, and Aaron Courville.
\newblock Mutual information neural estimation.
\newblock In {\em International Conference on Machine Learning}, pages
  530--539, 2018.

\bibitem{goyal2019infobot}
Anirudh Goyal, Riashat Islam, Daniel Strouse, Zafarali Ahmed, Matthew
  Botvinick, Hugo Larochelle, Yoshua Bengio, and Sergey Levine.
\newblock Infobot: Transfer and exploration via the information bottleneck.
\newblock {\em ICLR2019}, 2019.

\bibitem{puterman1994markov}
Martin~L Puterman.
\newblock Markov decision processes: Discrete stochastic dynamic programming.
\newblock 1994.

\bibitem{cover2012elements}
Thomas~M Cover and Joy~A Thomas.
\newblock {\em Elements of information theory}.
\newblock John Wiley \& Sons, 2012.

\bibitem{strouse2018learning}
DJ~Strouse, Max Kleiman-Weiner, Josh Tenenbaum, Matt Botvinick, and David~J
  Schwab.
\newblock Learning to share and hide intentions using information
  regularization.
\newblock In {\em Advances in Neural Information Processing Systems}, pages
  10249--10259, 2018.

\bibitem{gym_minigrid}
Maxime Chevalier-Boisvert, Lucas Willems, and Suman Pal.
\newblock Minimalistic gridworld environment for openai gym.
\newblock \url{https://github.com/maximecb/gym-minigrid}, 2018.

\bibitem{packer2018assessing}
Charles Packer, Katelyn Gao, Jernej Kos, Philipp Kr{\"a}henb{\"u}hl, Vladlen
  Koltun, and Dawn Song.
\newblock Assessing generalization in deep reinforcement learning.
\newblock {\em arXiv preprint arXiv:1810.12282}, 2018.

\bibitem{haarnoja2018soft}
Tuomas Haarnoja, Aurick Zhou, Pieter Abbeel, and Sergey Levine.
\newblock Soft actor-critic: Off-policy maximum entropy deep reinforcement
  learning with a stochastic actor.
\newblock {\em arXiv preprint arXiv:1801.01290}, 2018.

\bibitem{ppo}
Ilya Kostrikov.
\newblock Pytorch implementations of reinforcement learning algorithms.
\newblock \url{https://github. com/ikostrikov/pytorch-a2c-ppo-acktr, 2018.},
  2018.

\bibitem{haarnoja2017reinforcement}
Tuomas Haarnoja, Haoran Tang, Pieter Abbeel, and Sergey Levine.
\newblock Reinforcement learning with deep energy-based policies.
\newblock In {\em Proceedings of the 34th International Conference on Machine
  Learning-Volume 70}, pages 1352--1361. JMLR. org, 2017.

\end{thebibliography}
\clearpage
\section{Appendix}\label{appenndix}

\subsection{Proof for Lower Bound on Mutual Information by Variational Approximator}

This achieves an upper bound on $I(Z, S)$:
\begin{align*}
&\mathbb{E}_{S}[\KL\,[p(Z \vert S) \, \vert \, q(Z)]] \\
= &\int_{s}dx\,p(s)\int_{z}dz\,p(z \vert s)\,\log\frac{p(z\vert s)}{q(z)}\\
=& \int_{z, s}dx\,dz\,p(z \vert s)\, \log p(z \vert s) - \int_{z}dz\,p(z) \log q(z) \\
\geq& \int_{z, s}dx\,dz\,p(z \vert s)\, \log p(z \vert s) - \int_{z}dz\,p(z) \log p(z) \\
=&\int_{s}dx\,p(s)\int_{z}dz\,p(z \vert s)\,\log\frac{p(z\vert s)}{p(z)}\\
=&\; I(Z, S)
\end{align*}
where the inequality arises because of the non-negativeness KL-divergence: 
\begin{align}
\KL[p(z)\, \vert q(z)] &\geq 0 \\
\int_{z}dz\,p(z)\,\log p(z) &\geq \int_{z}dz\,p(z)\,\log q(z)
\end{align}

\subsection{Environment Descriptions}

\subsubsection{GridWorld}

The agent is a point that can move horizontally or vertically in a 2-D maze structure. Each state observation is a compact encoding of the maze, with each layer containing information about the placement of the walls, the goal position, and the agent position respectively. The goal state is one in which the goal position and the agent position are the same. The agent obtains a positive reward of $1$ when it reaches the goal, and no reward otherwise. 

\subsubsection{CartPole}

The agent is a cart sliding on a frictionless horizontal surface with a pole attached to its top. The pole is free to swing about the cart, and at each time step the cart moves to the left or to the right to keep the pole in upright position. Each sate observation consists of four variables: the cart position, the cart velocity, the pole angle, and the pole velocity at tip. The reward at every time $t$ is 1, and the episode terminates when it reaches 200 in length or when the pole fails to maintain an upright angle of at most $12$ degrees.

\subsubsection{HalfCheetah}

The agent is a bipedal robot with 6 joints and 8 links imitating a 2D cheetah. The agent moves horizontally on a smooth surface, and its goal is to learn to move in the positive direction without falling over, by applying continuous forces to each individual joint. The state observations encode the robot's position, velocity, joint angles, and joint angular velocities. The reward $r_t$ at each time $t$ is the robot's velocity in the positive direction, $v_{t} = x_{t} - x_{t-1}$, minus the action costs $\alpha\left\Vert a_t \right\Vert$. Here, $x_t$ indicates the position of the robot at time $t$, and $a_t$ is the robot's action input. 

\subsubsection{Humanoid}

The agent is a human-like robot with 13 rigid links and 17 actuators. The agent moves freely on a smooth surface, and its goal is to move in the forward direction as quickly as possible. Similar to HalfCheetah, its actions are continuous forces to each individual joint, and the state observations encoder its position, velocity, joint angles, and joint angular velocities. The reward at each time is the sum of its velocity ($v_{t} = x_{t} - x_{t-1}$) in the positive direction minus the action cost $\alpha\left\Vert a_t \right\Vert$.

\begin{table}
    \centering
    \begin{tabular}{c|cccc}
        Environment & Gridworld & CartPole & HalfCheetah & Humanoid\\
        State Dimensions & (12, 12, 3) & (4,) & (18,) & (47,)\\
        Action Dimensions & (4,)  & (2, ) & (6,) & (17, )\\
        Maximum Steps & 100 & 200 & 1000 & 1000
    \end{tabular}
    \caption{Environment dimensions and horizons}
    \label{tab:envdim}
\end{table}

\subsection {Network Parameters and Hyperparameters for Learning}

For all maze experiments we use standard A2C, and for all control experiments we use PPO. Our baseline is adopted from \cite{ppo}, and we modify the code to add a stochastic encoder.

\subsubsection{GridWorld}
For baseline, we use 3 layers of convolutional layers with 2-by-2 kernels, and channel size 16, 32, 64 respectively. The convolutional layers are followed by a linear layer ("deterministic encoder") of hidden size 64. Finally, the actor and critic each uses 1 linear layers of hidden size 64. We use Tanh activations between layers. For our approach, we add an additional linear layer after the convolution to output the diagonal variance of the encoder to provide stochasticity.

\subsubsection{CartPole}

For baseline, we use 1 linear layer of hidden size 32, followed by an additional linear layer of hidden size 32 ("deterministic encoder"). Actor and critic each uses 2 linear layers of hidden size 32. For our approach, we again add an additional linear layer of hidden size 32 after the first linear layer to output the diagonal variance for the stochastic encoder.

\subsubsection{HalfCheetah}

We follow mostly the same architecture as for CartPole, except the hidden size is 128.

\subsubsection{Humanoid}

For baseline, we use 2 linear layers of hidden size 96, followed by an additional linear layer of hidden size 96 ("deterministic encoder"). Actor and critic each uses 2 linear layers of hidden size 96. For our approach, we add an additional linear layer of hidden size 96 after the first 2 linear layers to output the diagonal variance. 

\subsection{Hyperparameter Selection}

The most crucial hyperparameter value is $\beta$, which determines the size of the information bottleneck. We evaluate the policy at even intervals during annealing to find optimal representations and control policies for each $\beta$ to determine the optimal $\beta$ value. For all other hyperparameters, we mostly followed the hyperparameters used in each environment's respective baselines, with the exception of tuning the learning rates, batch size, and encoder dimension. Learning rate was tuned through random initialization and short training; batch size and encoder dimension were turned through a binary sweep.

We provide hyperparameter choices in Table \ref{tab:gridhyp}, Table \ref{tab:reahyp}, and Table \ref{tab:humhyp} respectively.

\begin{table}
    \centering
    \begin{tabular}{c|c}
        Parameter    & Value\\
    gamma & 0.99 \\
    entropy coefficient & 0.01 \\
    leaning rate & $7 \times 10^{-4}$ \\
    gae-lambda coef & 0.95 \\ 
    value loss coef & 0.5 \\ 
    encoder dimension & 64 \\
    $\beta$ & 0.005
    \end{tabular}
    \caption{Hyperparameters for GridWorld}\label{tab:gridhyp}
\end{table}

\begin{table}
    \centering
    \begin{tabular}{c|c}
        Parameter    & Value\\
    gamma & 0.99 \\
    entropy coefficient & 0.0 \\
    leaning rate & $3 \times 10^{-4}$ \\
    clip range & $[-0.2, 0.2]$ \\
    max gradient norm & 0.5\\
    batch size & 128 \\
    gae-lambda coef & 0.95 \\ 
    entropy coef & 0.01 \\
    value loss coef & 0.5 \\ 
    
    CartPole encoder dimension & 32 \\
    HalfCheetah encoder dimension & 128 \\
    $\beta$ for CartPole & 5e-5\\
    $\beta$ for HalfCheetah & 5e-4\\
    \end{tabular}
    \caption{Hyperparameters for CartPole, HalfCheetah}\label{tab:reahyp}
\end{table}

\begin{table}
    \centering
    \begin{tabular}{c|c}
        Parameter    & Value\\
    gamma & 0.99 \\
    entropy coefficient & 0.0 \\
    leaning rate & $5 \times 10^{-6}$ \\
    clip range & $[-0.2, 0.2]$ \\
    max gradient norm & 0.5\\
    batch size & 128 \\
    gae-lambda coef & 0.95 \\ 
    entropy coef & 0 \\
    value loss coef & 1 \\ 
    
    encoder dimension & 96 \\
    $\beta$ & 8e-3
    \end{tabular}
    \caption{Hyperparameters for Humanoid}\label{tab:humhyp}
\end{table}

\subsection{Evaluation Results for Extreme Configurations in Cartpole}

\begin{figure}[h]
    \centering
    \includegraphics[width=0.45\textwidth]{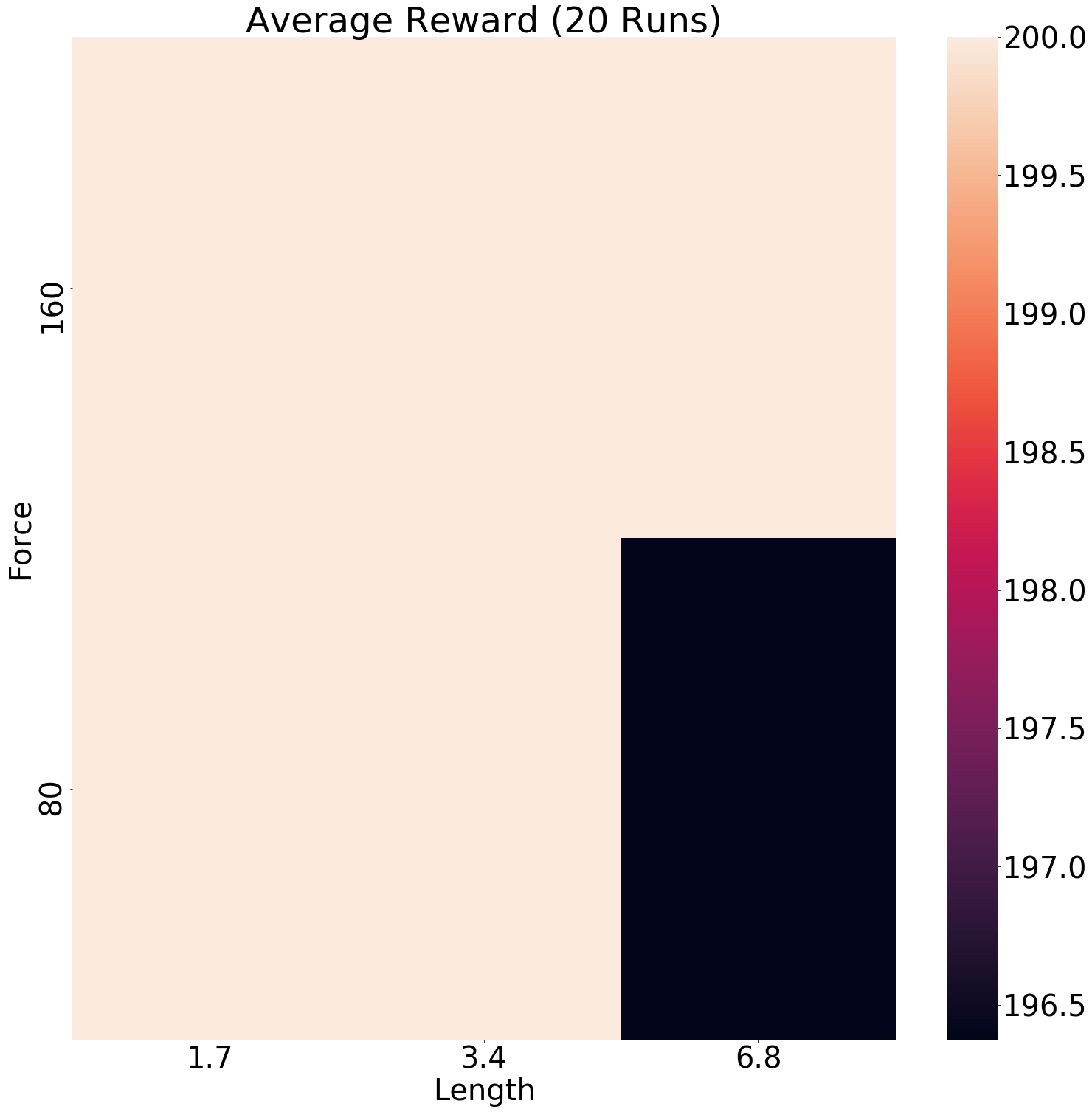}
    \caption{Evaluation performance of policy with information bottleneck on extreme configurations in CartPole. The x-axis indicates the length of the pole, while the y-axis indicates the push force of the cart. Each evaluation result is averaged over 20 episodes. Note that the agent achieves near-optimal reward in all 6 configurations. }
    \label{img:cp_extreme}
\end{figure}

We provide the evaluation grid for extreme configurations in Cartpole in Figure \ref{img:cp_extreme}.
\subsection{Full Evaluation Results Along Annealing Curve for Cartpole and HalfCheetah}

\begin{figure}
    \centering
    \includegraphics[width=1\textwidth]{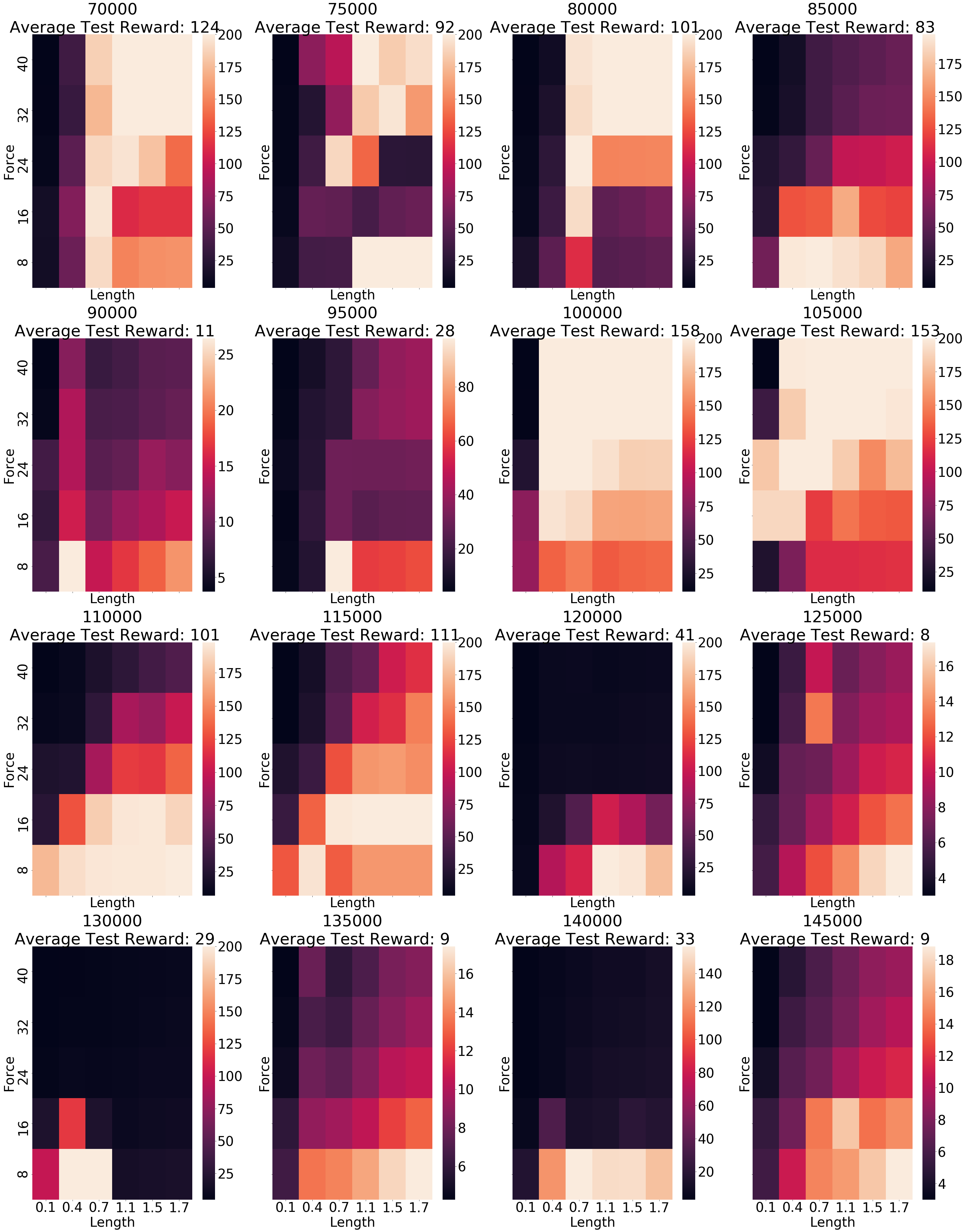}
    \caption{Full evaluation results for CartPole policies trained with an information bottleneck through annealing. Each subplot's title indicates the iteration number, the x-axis the pole length, the y-axis the push force, and the value of each cell the evaluation reward averaged over 20 episodes. The best policy was found at iteration 10,000, which corresponds to a $\beta$ value of 5e-5}
    \label{img:cp_full}
\end{figure}

\begin{figure}
    \centering
    \includegraphics[width=1\textwidth]{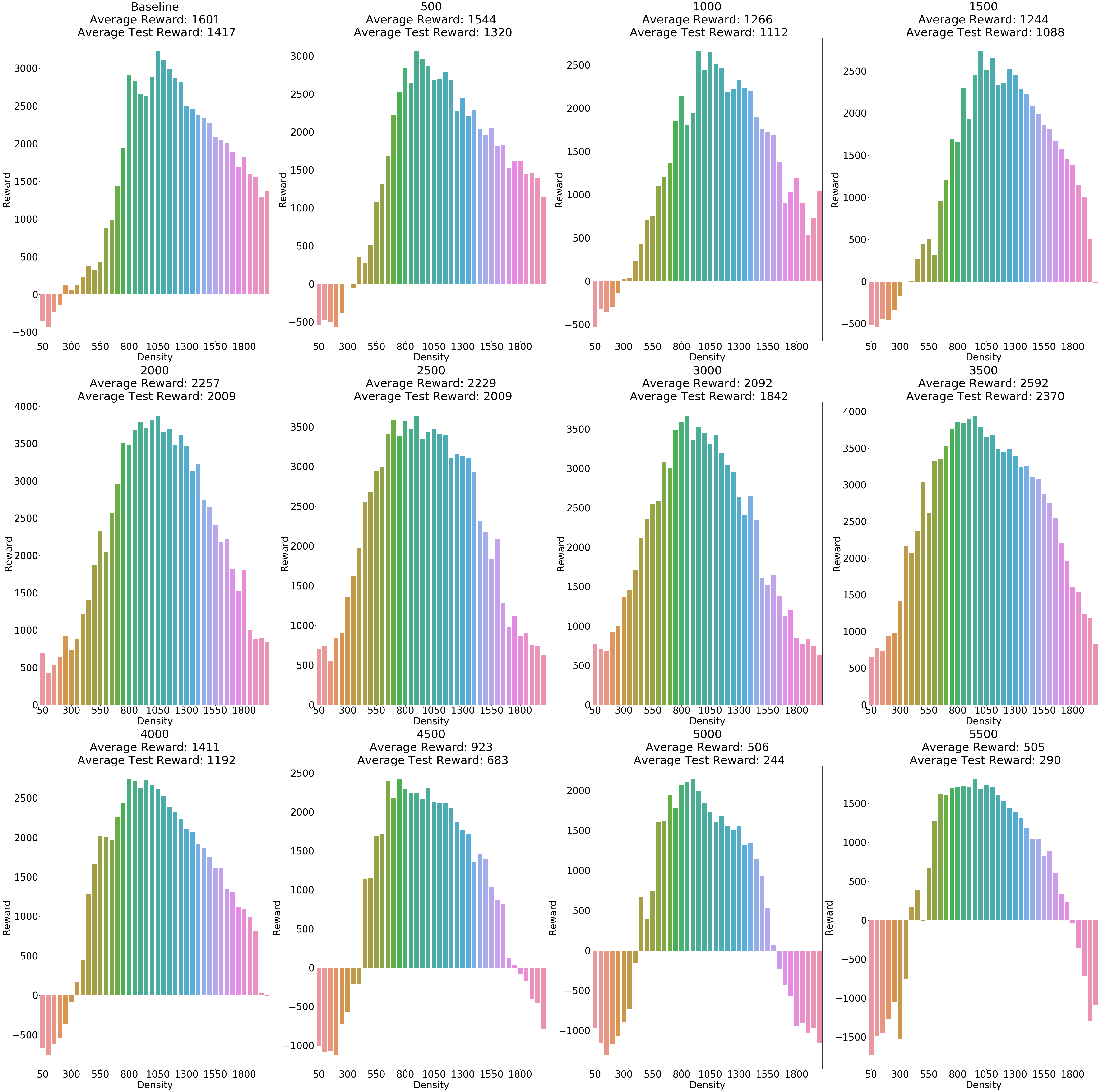}
    \caption{Full evaluation results for HalfCheetah policies trained with an information bottleneck through annealing. Each subplot's title indicates the iteration number, overall average reward across all configurations and average test reward across all test configurations. The x-axis indicates the robot's torso length, and the y-axis indicates the evaluation reward averaged over 20 episodes. The best policy was found at iteration 3500, which corresponds to a $\beta$ value of 5e-4.}
    \label{img:hc_full}
\end{figure}

We provide the full evaluation results for CartPole and HalfCheetah along their respective annealing curves in Figure \ref{img:cp_full} and Figure \ref{img:hc_full}. For each set of plots, we demonstrate the increase in generalization performance due to tightening of the information bottleneck, followed by a sudden deterioration of the policy as the encoder loses too much information.


\end{document}